\documentclass[lettersize,journal]{IEEEtran}
\usepackage{amsmath,amsfonts}
\usepackage{algorithmic}
\usepackage{algorithm}
\usepackage{array}
\usepackage[caption=false,font=normalsize,labelfont=sf,textfont=sf]{subfig}
\usepackage{textcomp}
\usepackage{stfloats}
\usepackage{url}
\usepackage{verbatim}
\usepackage{cite}
\usepackage[pdftex]{graphicx}
\usepackage{booktabs}
\usepackage{multirow}
\usepackage{tabularx}
\usepackage{CJKutf8} 
\usepackage{hyperref}
\usepackage{orcidlink}
\usepackage{xcolor}

\begin{document}

\title{Robust Dynamic Facial Expression Recognition}

\author{Feng~Liu$^{\orcidlink{0000-0002-5289-5761}}$,~\IEEEmembership{Member,~IEEE,}
        Hanyang~Wang$^{\orcidlink{0000-0002-5227-5765}}$, Siyuan~Shen$^{\orcidlink{0000-0001-9732-9600}}$
\thanks{Feng Liu is with School of Psychology, Shanghai Jiao Tong University, China, 200030. Hanyang Wang is with Midea Group Inc., Shanghai, China. Siyuan Shen is with Baidu Inc., Shanghai, China.

\protect
\IEEEcompsocthanksitem Feng~Liu(lsttoy@163.com / liu.feng@sjtu.edu.cn) is with the corresponding author.


}
\thanks{Manuscript received Nov. 27, 2024; revised Feb. 21, 2025.}}

\markboth{IEEE Transactions on Biometrics, Behavior, and Identity Science,~Vol.~XX, No.~X, XX~2025}%
{Shell \MakeLowercase{\textit{Feng Liu et al.}}:Robust Dynamic Facial Expression Recognition}
\maketitle


\begin{abstract}
The study of Dynamic Facial Expression Recognition (DFER) is a nascent field of research that involves the automated recognition of facial expressions in video data. Although existing research has primarily focused on learning representations under noisy and hard samples, the issue of the coexistence of both types of samples remains unresolved. In order to overcome this challenge, this paper proposes a robust method of distinguishing between hard and noisy samples. This is achieved by evaluating the prediction agreement of the model on different sampled clips of the video. Subsequently, methodologies that reinforce the learning of hard samples and mitigate the impact of noisy samples can be employed. Moreover, to identify the principal expression in a video and enhance the model's capacity for representation learning, comprising a key expression re-sampling framework and a dual-stream hierarchical network is proposed, namely Robust Dynamic Facial Expression Recognition (RDFER). The key expression re-sampling framework is designed to identify the key expression, thereby mitigating the potential confusion caused by non-target expressions. RDFER employs two sequence models with the objective of disentangling short-term facial movements and long-term emotional changes. The proposed method has been shown to outperform current State-Of-The-Art approaches in DFER through extensive experimentation on benchmark datasets such as DFEW and FERV39K. A comprehensive analysis provides valuable insights and observations regarding the proposed agreement. This work has significant implications for the field of dynamic facial expression recognition and promotes the further development of the field of noise-consistent robust learning in dynamic facial expression recognition. The code is available from [https://github.com/Cross-Innovation-Lab/RDFER].
\end{abstract}


\section{Introduction}
\IEEEPARstart{T}{he} role of facial expressions in human communication has been extensively studied \cite{wang2022systematic,liu2022opo,liu2022evogan, 10.1145/3242969.3264992, 10.1145/3503161.3547754, 10.1145/3503161.3547936, 10.1145/3551876.3554813, 10.1145/3503161.3548243, 2023Facial, Liu2024}. Understanding emotional states through facial expressions is crucial in social interactions, making automatic recognition of such expressions a critical challenge in various fields such as speech emotion recognition \cite{doi:10.34133/icomputing.0073}, mental health diagnosis \cite{lautman2022use，LIU2024101452}, driver fatigue monitoring \cite{li2021novel}, and metahuman technology \cite{fang2021metahuman}. While impressive progress has been made in Facial Expression Recognition (FER) \cite{wang2020suppressing,li2022towards}, the recognition of Dynamic Facial Expressions (DFER) is gaining increasing attention.

\begin{figure}[ht!]
  \centering
  \includegraphics[width=0.9\linewidth]{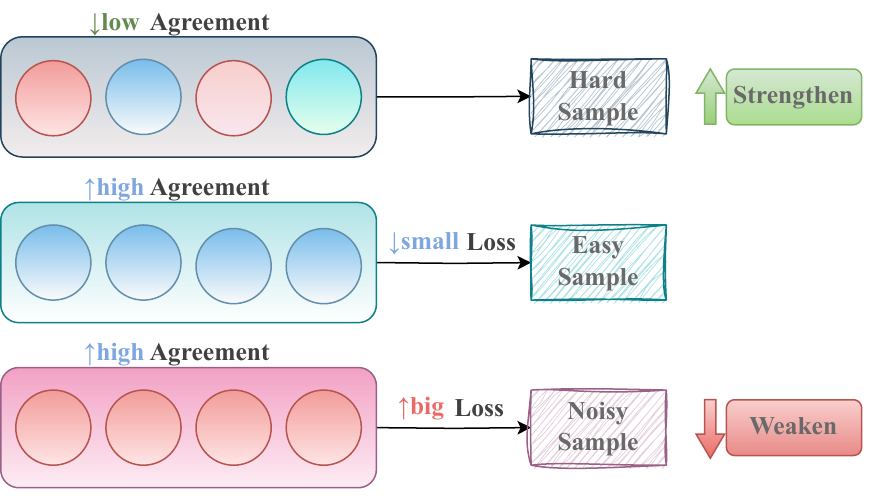}

  \caption{The proposed methodology for discriminating between hard and noisy samples involves segmenting the video into clips. If the clips are classified as distinct categories (having circles with different colors), the video is deemed challenging, and the model is reinforced in its learning. Conversely, if the clips are classified as the same category (having circles with the same color) but possess a significant loss, the video is regarded as noisy, and the model is prevented from learning.}
  \label{fig:cover}
\end{figure}

The emergence of extensive large-scale in-the-wild datasets, such as DFEW \cite{jiang2020dfew} and FERV39K \cite{wang2022ferv39k}, has led to the development of several methods \cite{zhao2021former,ma2022spatio,li2022nr,li2022intensity,wang2022dpcnet} for DFER. Previous studies \cite{zhao2021former,ma2022spatio} have primarily utilized general video understanding techniques for dynamic facial expression recognition. Later on, Zhang et al. \cite{10.1007/978-3-031-19809-0_24} propose an erasing attention consistency method to suppress the noisy samples during the training process automatically, Li et al. \cite{li2022nr} identify a significant amount of noisy frames in DFER and Li et al. \cite{li2022intensity} recognize that the expressions in DFER exhibit considerable intra-class and minimal inter-class differences. Some studies have also explored the physical structure of the face\cite{Zhang2017FacialER,NEURIPS2021_9332c513} and the movement of facial muscles\cite{Wang2023HTNetFM} from a number of perspectives. In addition, in our previous work M3DFEL\cite{wang2023rethinking}, we also significantly improved dynamic facial expression recognition by utilizing multi-example learning for Bag-level design against noise in dynamic expression recognition. However, we argue that some reported task-specific issues can be formed as general issues, for example, the weakly supervised problem contains the noisy frame problem, and face recognition also faces the problem of large intra-class variance and small inter-class variance. As shown in Fig. \ref{fig:cover}, in this study, we explore the datasets and formalize the hard and noisy sample issue present in DFER.

The issue of noisy labels is prevalent in most crowd-sourced datasets. The employment of the 'small loss trick' is a common approach to mitigate the effects of label noise. This approach considers samples with low loss values as clean and employs them to train the model. Several techniques such as loss reweighting\cite{9325065,10094303}, label refurbishing\cite{10516662}, sample selection\cite{10443531} and co-teaching have been developed based on this approach. However, the noisy label problem in DFER is more complex as the high-loss samples involves not only noisy samples, but also hard samples. Avoiding the learning of noisy labels would involve discarding hard samples, while mining hard samples may result in learning noisy labels. Our experimental results in Table \ref{tab:ablat_loss} demonstrate that both decreasing and increasing the loss of high-loss samples adversely impact performance.

Several recent studies have reported the issue of noisy labels in the long-tail scenario, where both small-class samples and noisy label samples incur significant losses, and is similar to the problem in DFER. These studies have employed various methods, such as self-supervised or meta-learning techniques, including contrastive learning \cite{karthik2021learning} and prototype learning \cite{wei2021robust} to mitigate the impact of noisy labels, or training a specialized classifier to convert hard noises into easy ones \cite{yi2022identifying}. While learning with fewer labels can enhance the robustness of the model against noise, utilizing the complete set of labels can further enhance performance. Training the classifier requires generating multiple predefined additional environments, and this process must be repeated several times. Therefore, an effective and efficient approach to differentiate between noisy samples and hard samples is needed. 

This paper presents a novel agreement-based approach for distinguishing between hard and noisy samples in the domain of DFER. It introduces a key expression re-sampling framework and a dual-stream hierarchical network (RDFER) that will revolutionise the way we approach this problem. Empirically, hard samples in DFER are often characterized by complex facial movements, ambiguous expressions, and environmental distractions, leading to varying predictions across different frames. Conversely, a noisy sample confounds the model when it is easy to predict but incurs a large loss. To address these issues, we propose utilizing the degree of agreement in the model's prediction across different sampled clips of a video to determine whether it is noisy or hard, and adjust its loss accordingly. Additionally, we introduce a Key Expression Re-sampling Framework and a Dual-Stream Hierarchical Network to identify key expressions in a video and enhance the model's representation learning abilities. Specifically, our re-sampling framework utilizes a small network to rapidly scan the entire video and identify the key expression, followed by the Dual-Stream Hierarchical Network to recognize the expression from the re-sampled key expression. The Dual-Stream Hierarchical Network leverages two sequence models in a hierarchical manner to disentangle short-term facial movements and long-term emotional changes respectively. We conduct extensive experiments to demonstrate the efficacy of our approach and provide further insights and findings through thorough analysis. Our numerical results indicate that our method achieves state-of-the-art performance.

Overall, our contributions can be summarized as follows:
\begin{itemize}
    \item We propose RDFER, an innovative agreement-based approach that enables the identification of hard and noisy samples in DFER and adjusts the loss function accordingly.
    \item We introduce the key expression re-sampling framework and the dual-stream hierarchical network, which serve to enhance the model's capacity to identify key expressions and reinforce the process of representation learning by disentangling short-term movements and long-term emotional changes.
    \item We conduct a series of comprehensive experiments to validate the efficacy of our proposed SOTA method, and present a detailed analysis that offers valuable insights and observations.
    \item Our study is of the utmost importance, as it concerns the advancement of dynamic facial expression recognition techniques and may pioneer the field of robust learning for dynamic facial expression recognition.
\end{itemize}

\section{Related Work}
\label{sec:relat}

\textbf{Dynamic Facial Expression Recognition.} In contrast to SFER techniques that focus solely on spatial features within an image, DFER approaches must consider both spatial and temporal information simultaneously. Transformer-based networks have gained popularity in recent literature for their ability to extract spatial and temporal information. Zhao et al. \cite{zhao2021former} introduce the dynamic facial expression recognition transformer (Former-DFER), comprising a convolutional spatial transformer (CS-Former) and a temporal transformer (T-Former). Ma et al. \cite{ma2022spatio} propose the spatial-temporal Transformer (STT) to capture discriminative features within each frame and model contextual relationships among frames. To reduce the impact of noisy frames on the dynamic facial expression recognition (DFER) task, a dynamic-static fusion module is used to extract more robust and discriminative spatial features from both static and dynamic features \cite{li2022nr,li2022intensity}. Wang et al. \cite{wang2022dpcnet} propose the Dual Path multi-excitation Collaborative Network (DPCNet) to learn critical information for facial expression representation from fewer key frames in videos. In addition, Zhang et al. proposed to try to improve dynamic expression recognition by means of increasing zero-shot generalization ability\cite{zhang2025generalizable}, sample feature diversification\cite{zhang2024leave} and symmetric noisy labels\cite{zhang2024open}.


\begin{figure*}[ht!]
  \centering
  \includegraphics[width=0.9\linewidth]{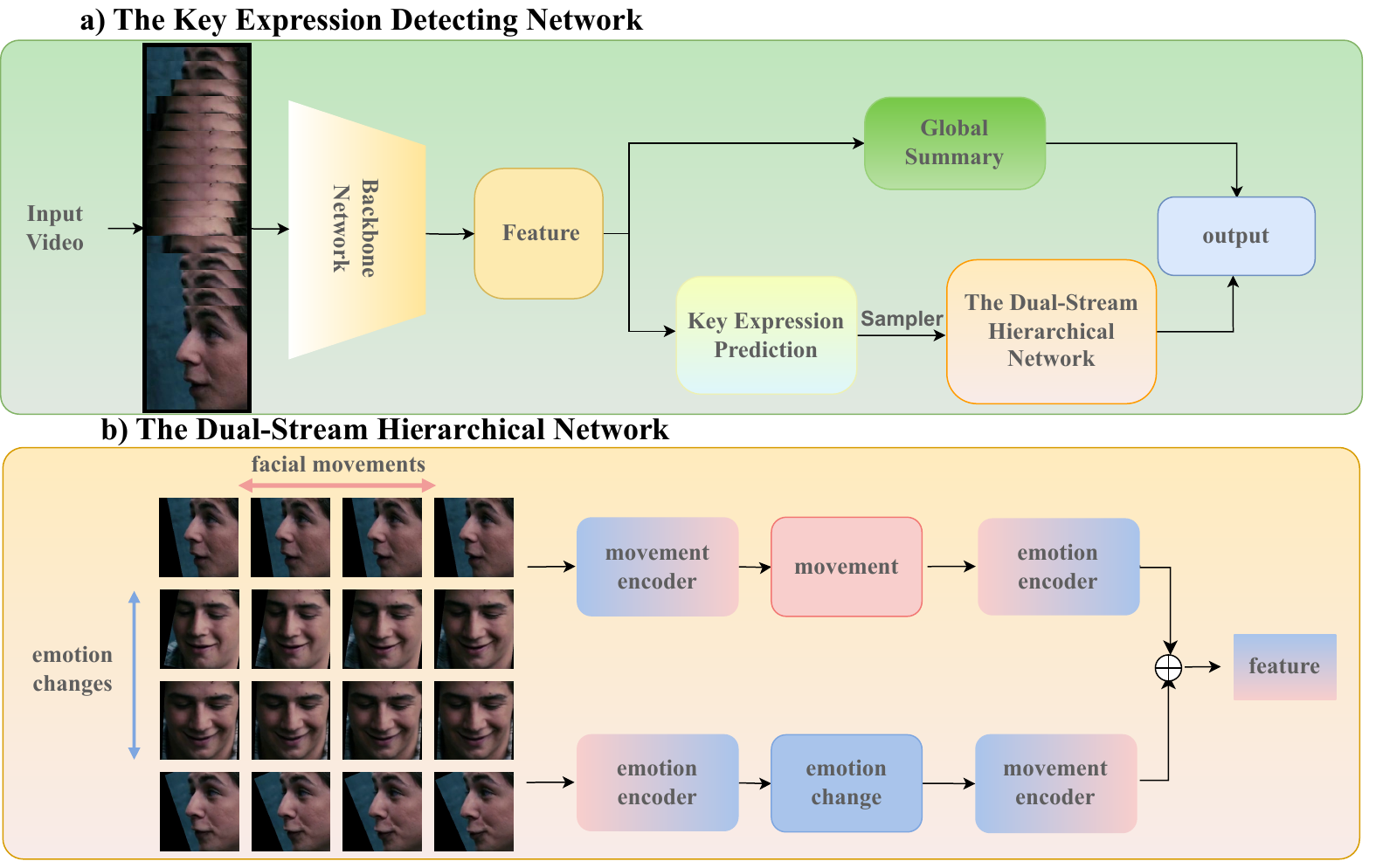}

  \caption{An overview of the Key Expression Re-sampling Framework. (a) The Key Expression Detecting Network. The input video is sampled uniformly and fed into a tiny backbone network to quickly obtain a global summary and predict the key expression. (b) The Dual-Stream Hierarchical Network. Taken the key expression predicted by (a), this network learns the representation through disentangling the short-term facial movements and long-term emotion changes with a dual-stream hierarchical design. }
  \label{fig:architecture}
\end{figure*}

\textbf{Long-tailed noisy label problem.} The topic of long-tailed noisy label problem has gained prominence due to its relevance to real-world applications and its inherent complexity. The effectiveness of existing noisy label methods, which rely on small loss tricks, has been found to be limited when dealing with long-tail situations. Therefore, there is a need for a new framework to address this challenge. RoLT \cite{wei2021robust} has identified this issue and proposed a prototypical noise detection method that is robust to label noise by designing a distance-based metric. JA+SL \cite{karthik2021learning} has demonstrated that self-supervised learning approaches are effective in handling severe class imbalance. Additionally, H2E \cite{yi2022identifying} has defined long-tailed noises as 'hard' noises and proposed the Hard-to-Easy framework to convert 'hard' noises to 'easy' ones.


\section{Methodology}
\label{sec:metho}

\subsection{The Agreement-Based Loss Reweighting}
Inspired by decoupling \cite{malach2017decoupling}, which leverages the disagreement between two networks to discern noisy labels, our aim is to employ the discordance of a single network across distinct frames to differentiate between samples. Given that the videos in DFER datasets are limited to a few seconds, we anticipate that facial movements between frames will be subtle and unlikely to have a significant impact on emotion recognition results. Nevertheless, much like how identical networks with different initializations can exhibit disagreement on the same input, a network may exhibit discordance on inputs with minor variations. By assessing such disagreement, we can determine whether a given sample is challenging or not.

As previously mentioned, the presence of hard and noisy samples presents a challenge in distinguishing between the two using the small loss trick. Consequently, the model must concentrate on learning from the hard samples while disregarding the noisy labels. In the context of DFER, the factors that contribute to the difficulty of a sample include complex facial movements, ambiguous expressions, and environmental distractions. In addition to these obstacles, temporal changes in the samples pose a significant challenge in comparison to those in SFER. The degree of disagreement between the temporal changes and the expression recognition results serves as a direct indicator of the sample's difficulty. Notably, facial movements typically do not have a substantial impact on the final outcome. Therefore, a strong disagreement is a clear indication of a hard sample.

To put this formally, given a video $V \in R^{T \times C \times H \times W}$ and split it into frames or clips $V_i$. By using a recognizer, we obtain the facial expression of them,
\begin{equation}
E_i = f(V_i). 
\end{equation}
The agreement $A$ can be calculated by counting the proportion of the most predicted class. For example, when four clips sample from one video are predicted as \textit{Neutral}, \textit{Surprise}, \textit{Happy} and \textit{Happy}, the video is considered having two \textit{Happy}s in four clips and the corresponding agreement is 0.5.

Our implementation employs fundamental techniques to establish and manage hard and noisy samples, thus enabling the straightforward observation of the agreement's impact. A hard sample is defined as a sample with minimal agreement, while a falsely labeled sample demonstrates substantial agreement but significant loss. To this end, we ascribe two ranks to all samples based on their agreement and loss, and subsequently set two thresholds. Specifically, the $t_h$ percent of samples with the minimum consistency are considered as hard samples, the $t_n$ percent of samples with the maximum agreement and maximum loss are considered as noisy samples. Any samples falling outside these thresholds are deemed ordinary. Consequently, the loss calculation can be derived using the following equation:

\begin{equation}
    loss = \left \{
    \begin{array}{lr}
         \lambda _{noisy} \times Cross\_Entorpy(x, y) & x \in noisy \\
         \lambda _{ordinary} \times Cross\_Entorpy(x, y) & x \in ordinary \\
         \lambda _{hard} \times Cross\_Entorpy(x, y) & x \in hard
    \end{array}
\right.
\end{equation}

\subsection{The Key Expression Re-sampling Framework}

Due to the presence of irrelevant movements in a proportion of the samples, the model may become confused. To address this issue, we propose the Key Expression Re-sampling Framework to first locate the position of the key expression, thereby excluding irrelevant frames. Fig. \ref{fig:architecture} illustrates the architecture employed. Initially, the input frames are uniformly sampled and instantiated by frame to generate the location parameters for key expressions by the detecting network. Subsequently, the Dual-Stream Hierarchical Network extracts the discriminative facial movements in the re-sampled key expressions and predicts the expression. For instance, in a video $V$ with 64 frames, we first uniformly sample $[V_0, V_4, V_8, ..., V_{60}]$ as the input of the detecting network. The output location parameters of $\mu, \delta$ correspond to the re-sampled frames of $[...,V_{64\times \mu -\delta}, V_{64\times \mu},V_{64\times \mu +\delta},...]$. 

The detecting network comprises the backbone, the key expression detecting module, and the summary module, as illustrated in Fig. \ref{fig:architecture}. The backbone network can be any 2D/3D CNN/ViT Networks for video recognition and extract the feature,
\begin{equation}
    f = F_b(V),
\end{equation}
where variables $f$ and $F_b$ are used to denote the output feature and backbone network, respectively. 

The key expression detection module utilizes the feature map as input to generate positional parameters, denoted as $\mu$ and $\delta$, representing center position and stride of the key expressions, for every sample.
\begin{equation}
  \mu, \delta = F_d(f).
\end{equation}
The fully connected layer $F_d$ is involved in this process. The re-sampled key expressions provide additional local information regarding facial movements; however, it is desirable to infer facial expressions using both local and global information. To address this issue, the detection network also produces a global summary of the video, which is subsequently passed to the Dual-Stream Hierarchical network,
\begin{equation}
  S = Sigmoid(F_s(f)),
\end{equation}
where $F_s$ is a fully connected layer.

\subsection{The Dual-Stream Hierarchical Network}

The architecture of the Dual-Stream Hierarchical Network is illustrated in Fig. \ref{fig:architecture}. The foundation of the design is predicated upon the notion that brief temporal correlations are indicative of minor facial movements, whereas long-term temporal correlations signify dynamic emotional alterations in facial expressions. These two temporal correlations possess distinct properties, with short-term facial movements primarily encompassing pixel-level modifications that convey the subject's expression. Conversely, long-term emotional changes entail high-level semantic modifications. Consequently, we utilize two sequence models to individually model these two temporal relationships and disentangle them, thereby enabling the acquisition of a superior representation.

To begin, we utilize a 2D image backbone to encode the frames into a spatial feature $F \in R^{n \times C}$, where $n$ represents the number of frames and $C$ represents the channel. The feature is then rearranged in two ways for the movement-emotion stream and the emotion-movement stream. The movement-emotion stream analyzes the movement in the cropped clips first, followed by the emotional change in the processed clips. Conversely, the emotion-movement stream analyzes the emotional change through frames with stride first, followed by the analyzing the movement change. The output feature is the concat of the two streams and followed by a fully connected layer to predict the final expression.

\begin{algorithm}[ht!]
    \caption{Pseudo code for Dual-Stream Hierarchical Network.}
    
\textbf{Input:} $R$: raw feature, $S_{m/e}$: Movement/Emotion Sequence Model, 
    
\textbf{Output:} $f$: feature
    
$R_{me} = rearrange(R, 'n C ->n_1 n_2 C')$

$R_{em} = rearrange(R, 'n C ->n_2 n_1 C')$
    
$f_{me} = S_e(S_m(R_{me}))$

$f_{em} = S_m(S_e(R_{em}))$

$f = f_{me} + f_{em}$

$ return$ $f$
    
    \label{algo:net}
\end{algorithm}

\begin{table*}
    \centering
    \begin{tabular}{c|ccccccc|cc} 
        \toprule
        \multirow{2}{*}{\textbf{Method}} & \multicolumn{7}{c|}{\textbf{Accuracy of Each Emotion↑(\%)}} & \multicolumn{2}{c}{\textbf{Metrics↑(\%)}} \\ \cline{2-10} 
         & \textbf{Hap.} & \textbf{Sad} & \textbf{Neu.} & \textbf{Ang.} & \textbf{Sur.} & \textbf{Dis.} & \textbf{Fea.} & \textbf{WAR} & \textbf{UAR} \\ 
        \midrule
        C3D\cite{tran2015learning} & 75.17 & 39.49 & 55.11 & 62.49 & 45.00 & 1.38 & 20.51 & 53.54 & 42.74 \\
        P3D\cite{qiu2017learning} & 74.85 & 43.40 & 54.18 & 60.42 & 50.99 & 0.69 & 23.28 & 54.47 & 43.97 \\
        I3D\cite{carreira2017quo} & 78.61 & 44.19 & 56.69 & 55.87 & 45.88 & 2.07 & 20.51 & 54.27 & 43.40 \\
        R(2+1)D18\cite{tran2018closer} & 79.67 & 39.07 & 57.66 & 50.39 & 48.26 & 3.45 & 21.06 & 53.22 & 42.79 \\
        3D ResNet18\cite{hara2018can} & 73.13 & 48.26 & 50.51 & 64.75 & 50.10 & 0.00 & 26.39 & 54.98 & 44.73 \\
        ResNet18+LSTM\cite{jiang2020dfew} & 78.00 & 40.65 & 53.77 & 56.83 & 45.00 & \textbf{4.14} & 21.62 & 53.08 & 42.86 \\
        EC-STFL\cite{jiang2020dfew} & 79.18 & 49.05 & 57.85 & 60.98 & 46.15 & 2.76 & 21.51 & 56.51 & 45.35 \\
        FormerDFER\cite{zhao2021former} & 84.05 & 62.57 & 67.52 & 70.03 & 56.43 & 3.45 & \underline{31.78} & 65.70 & 53.69 \\
        STT\cite{ma2022spatio} & 87.36 & 67.90 & 64.97 & 71.24 & 53.10 & \underline{3.49} & \textbf{34.04} & 66.45 & 54.58 \\
        DPCNet\cite{wang2022dpcnet} &-&-&-&-&-&-&- & 66.32 & 55.02*  \\
        NR-DFERNet\cite{li2022nr} & 88.47 & 64.84 & 70.03 & \underline{75.09} & 61.60 & 0.00 & 19.43 & 68.19 & 54.21\\ 
        GCA+IAL\cite{li2022intensity} & 87.95 & 67.21 & \underline{70.10} & \textbf{76.06} & \textbf{62.22} & 0.00 & 26.44 & 69.24 & 55.71\\ 
        M3DFEL\cite{wang2023rethinking} & \underline{89.59} & \underline{68.38} & 67.88 & 74.24 & 59.69 & 0.00 & 31.63 & \underline{69.25} & \underline{56.10}\\ 
        \midrule
        \textbf{RDFER(Ours)} & \textbf{89.69} & \textbf{69.22} & \textbf{70.18} & 71.47 & \underline{62.08} & 0.69 & 28.71 & \textbf{69.73} & \textbf{56.93} \\
        \bottomrule
    \end{tabular}
    \caption{Comparison(\%) of our Method with the state-of-the-art methods on DFEW. * indicates the result is calculated according to the confusion matrix reported in the paper. (Bold: Best result, Underline: Second best, the abbreviations represent: Happy, Sad, Neutral, Angry, Surprise, Disgust, Fear)}
    \label{tab:compa_dfew}
\end{table*}

\section{Experiments}

\subsection{Datasets and Metrics }
\label{sec:datas}

The DFEW dataset \cite{jiang2020dfew} is a large-scale collection of dynamic facial expression samples obtained from over 1,500 movies worldwide and comprises more than 16,000 video clips. Methods are evaluated using the 5-fold cross-validation setting provided by DFEW, with the weighted average recall (WAR) and unweighted average recall (UAR) used as evaluation metrics.

FERV39K \cite{wang2022ferv39k} is currently the largest in-the-wild DFER dataset, containing 38935 video clips collected from four scenarios and subdivided into 22 fine-grained scenes. The training and testing sets divided by FERV39k are directly used for a fair comparison.

\subsection{Implementation Details}
\label{sec:imple}

Our research framework is constructed using PyTorch, leveraging Tesla V100 GPUs. The backbone network and sequence model are comprised of the ResNet50 pretrained on ImageNet and BiLSTMs. A subset of 20\% of the samples are identified as hard samples, and their losses are augmented by a factor of 1.5. Additionally, 10\% of the samples are identified as noisy samples and are not included in the backward propagation process. The models are trained for 100 epochs, with ten warm-up epochs, using the AdamW optimizer and cosine scheduler. The learning rate is set to 7e-4, with a minimum learning rate of 7e-6, and a weight decay of 0.05. The batch size is set to 128, and label smoothing is utilized with a value of 0.1. Augmentation techniques utilized include random cropping, horizontal flipping, and 0.4 color jitter. The evaluation metrics employed are the weighted average recall (WAR) and unweighted average recall (UAR). Further experimentation and analysis utilizes DFEW \cite{jiang2020dfew}.

\subsection{Performance Comparison}
\label{sec:compa}

We compare our method with the state-of-the-art methods on two in-the-wild datasets DFEW and FERV39K. The WAR is used as a primary measure of accuracy to demonstrate the comprehensive performance of the method. Meanwhile, the UAR denotes the mean accuracy across all classes, providing insight into balanced performance among various categories.

In this study, the results on DFEW are presented in Table \ref{tab:compa_dfew}, where experiments are conducted using 5-fold cross-validation. Our proposed framework yield superior outcomes in both WAR and UAR, surpassing the performance of GCA+IAL \cite{li2022intensity} by 0.49\%/1.22\% for WAR/UAR. Additionally, Table \ref{tab:compa_dfew} provides the results for each expression, with a detailed analysis of the suboptimal performance of \textit{Disgust} and \textit{Fear} presented in Section \ref{sec:visua}.

\begin{table}[ht!]
  \centering
  \begin{tabular}{c|cc}
    \toprule
    \textbf{Method} & \textbf{WAR↑} & \textbf{UAR↑} \\
    \midrule
    C3D & 31.69\% & 22.68\% \\
    P3D & 33.39\% & 23.20\% \\
    I3D & 38.78\% & 30.17\% \\
    R(2+1)D18 & 41.28\% & 31.55\% \\
    3D ResNet18 & 37.57\% & 26.67\% \\
    R18+LSTM & 42.95\% & 30.92\% \\
    2R18+LSTM & 43.20\% & 31.28\%  \\
    FormerDFER & 45.72\% & \textbf{36.88}\%  \\
    NRDFERNet & 45.97\% & 33.99\% \\
    GCA+IAL & \underline{48.54}\% & 35.82\% \\
    M3DFEL & 47.67\% &  35.94\% \\
    \midrule
    \textbf{RDFER(Ours)} & \textbf{48.60}\% & \underline{36.47}\% \\
    \bottomrule
  \end{tabular}
  \caption{Comparison(\%) of our Method with the state-of-the-art methods on FERV39K.}
  \label{tab:compa_ferv39k}
\end{table}

Table \ref{tab:compa_ferv39k} presents the results for the FERV39K dataset, which is known for its complexity and lower accuracy compared to DFEW. Our proposed framework demonstrates superior performance when compared to 3DResNet18 \cite{hara2018can} and R18+LSTM \cite{wang2022ferv39k}, with improvements of 11.03\% and 5.65\% for WAR, and 9.80\% and 5.55\% for UAR, respectively, indicating its effectiveness. Furthermore, our approach outperforms the Transformer-based GCA+IAL \cite{li2022intensity}. It is noteworthy that FormerDFER stands out as a method that prioritizes UAR over WAR in FERV39K, resulting in higher accuracy for smaller classes such as \textit{Disgust} and \textit{Fear}, but lower overall performance.

\subsection{Ablation Study}
\label{sec:ablat}

\begin{table}[ht!]
  \centering
  \begin{tabular}{l|cc}
    \toprule
    \textbf{Configuration} & \textbf{WAR↑} & \textbf{UAR↑}\\
    \midrule
    \textbf{RDFER(Ours)}  & \textbf{69.73}\%  & \textbf{56.93}\% \\
    w/o Re-sampling & 68.16\%  & 55.89\% \\
    w/o Global Summary  & 69.31\%  & 56.71\% \\
    w/o Dual-stream Architecture  & 69.15\%  & 56.57\% \\
    w/o Hierarchical Architecture  & 69.22\%  & 56.69\% \\
    \bottomrule
  \end{tabular}
  \caption{The ablation Study of different network configurations.}
  \label{tab:ablat_arch}
\end{table}

\textbf{Evaluation of different network configurations.} The present study assesses various network configurations and reports the findings in Table \ref{tab:ablat_arch}. The elimination of the re-sampling framework results in the removal of the global summary. The ablation analysis shows that the re-sampling framework effectively detects key frames or clips and discards irrelevant frames, thereby enhancing performance by 1.57\%/1.04\% of WAR/UAR. Additionally, the global summary offers valuable insight into the context of the expression and enhances model performance. The ablation of the Dual-Stream Hierarchical Network involves using a single movement encoder and an emotion encoder consecutively, and results indicate that the dual-stream architecture facilitates better learning of short-term facial movements and long-term emotional changes. Moreover, the dual-stream hierarchical architecture outperforms the ensemble model of two sequence models without the hierarchical architecture, exhibiting an increase in performance by 0.51\%/0.24\% of WAR/UAR. Overall, the proposed architecture effectively disentangles short-term facial movements and long-term emotional changes, resulting in an improved representation of facial expressions.

\begin{table}[ht!]
  \centering
  \resizebox{\linewidth}{!}{
  \begin{tabular}{l|cc}
    \toprule
    \textbf{Configuration} & \textbf{WAR↑} & \textbf{UAR↑}\\
    \midrule
    vanilla Cross Entropy Loss  & 69.03\%  & 56.03\% \\ \midrule
    + big-loss strengthening & 68.83\%  & 55.77\% \\
    + big-loss weakening  & 68.63\%  & 55.73\% \\ \midrule
    + hard sample strengthening  & 69.35\%  & 56.51\% \\
    + noisy sample weakening  & 69.47\%  & 56.59\% \\
    + hard and noisy sample reweighting  & \textbf{69.73}\%  & \textbf{56.93}\% \\
    \bottomrule
  \end{tabular}}
  \caption{The ablation Study of different loss reweighting strategies.}
  \label{tab:ablat_loss}
\end{table}

\textbf{Evaluation of loss reweighting strategies.} The result are presented in Table \ref{tab:ablat_loss}. In our experimental approach, we aim to enhance or reduce the learning of high-loss samples through loss modification. However, the outcomes indicate a decrease in performance. The inadequacy of the small loss trick prompts us to explore an alternative approach. Consequently, we introduce the agreement metric, defining hard samples as those with low agreement and noisy samples as those with high agreement and high loss. The results demonstrate the efficacy of both strengthening the learning of hard samples and weakening the learning of noisy samples, indicating the ability of the agreement metric to differentiate between hard and noisy samples. By simultaneously reweighting the loss of challenging and erroneous samples, the model achieves a 0.70\%/0.90\% improvement in WAR/UAR.

\begin{table}[ht!]
    \centering
    \resizebox{\linewidth}{!}{
        \begin{tabular}{c|ccccccc|c}
        \toprule
        \textbf{Agr.} & \textbf{Ha.} & \textbf{Sa.} & \textbf{Ne.} & \textbf{An.} & \textbf{Su.} & \textbf{Di.} & \textbf{Fe.} & \textbf{Sum} \\ \midrule
        0.50 & 0.02 & 0.10 & 0.10 & 0.11 & 0.16 & 0.03 & 0.19 & 0.10 \\
        0.75 & 0.06 & 0.15 & 0.24 & 0.18 & 0.28 & 0.28 & 0.28 & 0.18 \\
        1.00 & 0.92 & 0.75 & 0.66 & 0.71 & 0.55 & 0.66 & 0.54 & 0.71 \\ \midrule
        Num. & 489 & 379 & 534 & 435 & 294 & 29 & 181 & 2341 \\
        \bottomrule
        \end{tabular}}
    \caption{The proportion of samples with different agreements and expressions.}
    \label{tab:ablat_agree_num}
\end{table}

\subsection{Effectiveness and Scalability}

\begin{table*}[ht]
    \centering
    \begin{tabular}{c|c|cc|cc|cc}
    \toprule
    Method & M3DFEL & WAR & $\Delta$ & UAR & $\Delta$ & Params(M) & FLOPs(G) \\ \midrule
    \multirow{2}{*}{CNN-LSTM}    & & 66.71\% & - & 55.38\% & - & 22.2 & 7.97  \\
                    & $\checkmark$ & 68.62\% & +1.91\% & 56.44\% & +1.06\% & 34.0 & 8.02\\ \hline
    \multirow{2}{*}{MC3}         & & 67.49\% & - & 56.10\% & - &  11.5 & 1.85 \\
                    & $\checkmark$ & 69.18\% & +1.69\% & 57.97\% & +1.87\% & 23.3 & 1.89\\ \hline
    \multirow{2}{*}{R(2+1)D}     & & 68.97\% & - & 56.69\% & - &  31.3 & 40.77 \\
                    & $\checkmark$ & 71.14\% & +2.17\% & 57.71\% & +1.02\% &  43.1 & 42.91 \\ \hline
    \multirow{2}{*}{R3D18}       & & 68.23\% & - & 55.36\% & - &  33.2 & 1.59\\
                    & $\checkmark$ & 70.40\% & +2.17\% & 56.81\% & +1.45\% & 45.0 & 1.65 \\ \hline
    \multirow{2}{*}{Former-DFER} & & 63.93\% & - & 52.70\% & - & 18.0 & 8.31 \\
                    & $\checkmark$ & 65.71\% & +1.78\% & 54.06\% & +1.36\% & 29.9 & 8.37 \\ \bottomrule
    \end{tabular}
    \caption{Ablation study of adopting different video backbones to RDFER.}
    \label{tab:backbone}
\end{table*}
The fundamental structure of RDFER is relatively straightforward and can be replicated with any video model as the underlying structure. Consequently, further studies were conducted to demonstrate the scalability of the proposed framework. Specifically, different video backbones were adopted for the framework, including R(2+1)D and MC3 provided by Torchvision, CNN-LSTM and the reproduced Former-DFER \cite{zhao2021former}. The experiments were conducted on the first fold of DFEW\cite{jiang2020dfew}.For each method, the first line represents setting the bag size to one and removing the Instance Aggregation Module, and the second line represents setting the bag size to four.During the adoption of Former-DFER, it was found to be sensitive, often resulting in training failure and slightly lower results. As demonstrated in Table \ref{tab:backbone}, the implementation of all methods has been observed to enhance WAR by approximately two points and UAR by one point within the designated framework. This outcome signifies the efficacy and scalability of the proposed framework.Notably, despite the augmented parameter count, the framework imposes a negligible computational cost of approximately 0.05G FLOPs. It is anticipated that subsequent research will enhance the framework in a manner analogous to the advancement in video understanding, SFER, and MIL.

\subsection{Analysis}
\label{sec:analy}

\textbf{The proportion of samples with different agreements and expressions.} The study is carried out using the first fold of DFEW. The findings are presented in Table \ref{tab:ablat_agree_num}. The video data are divided into four clips, and the level of agreement varied from one-quarter to four-quarters. The results indicate that the model was able to achieve agreement in the majority of the samples (71.46\%), with only 0.03\% of samples being predicted as four different types of clips. Furthermore, 18.24\% of samples have one clip that differed from the other clips. Additionally, different facial expressions exhibits varying degrees of agreement. Approximately 92\% of \textit{Happy} result in full agreement, while \textit{Sad}, \textit{Neutral}, \textit{Angry}, and \textit{Disgust} achieve agreement in around 65\% to 75\% of the samples. Only 55.10\% and 53.59\% of samples result in full agreement for \textit{Surprise} and \textit{Fear}, respectively. The results reveal notable differences in agreement among the facial expressions. Apart from the impact of class imbalance, classes with a comparable number of samples also display distinct agreement distributions. Specifically, \textit{Happy} have approximately 26\% more samples with full agreement than Neutral, implying that \textit{Neutral} is more perplexing than \textit{Happy}.

\begin{table}[ht!]
    \centering
    \resizebox{\linewidth}{!}{
        \begin{tabular}{c|ccccccc|c}
        \toprule
        \textbf{Agr.} & \textbf{Hap.} & \textbf{Sad.} & \textbf{Neu.} & \textbf{Ang.} & \textbf{Sur.} & \textbf{Dis.} & \textbf{Fea.} & \textbf{Ave.} \\ \midrule
        0.50 & 0.36 & 0.32 & 0.31 & 0.46 & 0.48 & 0.00 & 0.29 & 0.38 \\
        0.75 & 0.54 & 0.58 & 0.51 & 0.60 & 0.66 & 0.12 & 0.28 & 0.53 \\
        1.00 & 0.9 & 0.76 & 0.80 & 0.81 & 0.75 & 0.00 & 0.24 & 0.79 \\ \bottomrule
        \end{tabular}}
    \caption{The accuracy of samples with different agreements and expressions. (Agr. represents Agreement and Ave. represents Average.)}
    \label{tab:ablat_agree_acc}
\end{table}

\textbf{The accuracy of samples with different agreements and expressions.} The study is carried out on the first fold of the DFEW dataset, and the outcomes are presented in Table \ref{tab:ablat_agree_acc}. Notably, the findings indicate a conspicuous trend whereby samples with higher levels of agreement exhibit greater accuracy. This observation implies that agreement can serve as a reflection of sample difficulty, thereby supporting the notion that low/high agreement samples correspond to hard/easy samples. Similarly, Table \ref{tab:ablat_agree_num} demonstrates that, in terms of expressions, \textit{Happy} is the easiest expression to classify with high levels of agreement and accuracy. Conversely, \textit{Fear} and \textit{Disgust} exhibit low accuracy, with the former displaying approximately 55\% full agreement, but significantly lower accuracy than \textit{Surprise}. These results suggest that the confusion surrounding \textit{Fear} may be attributed to spatial rather than temporal issues.

\subsection{Visualization}
\label{sec:visua}

In order to conduct a more thorough evaluation of the efficacy of our approach, we undertake studies involving visualization.

\begin{figure*}[ht!]
  \centering
  \includegraphics[width=1\linewidth]{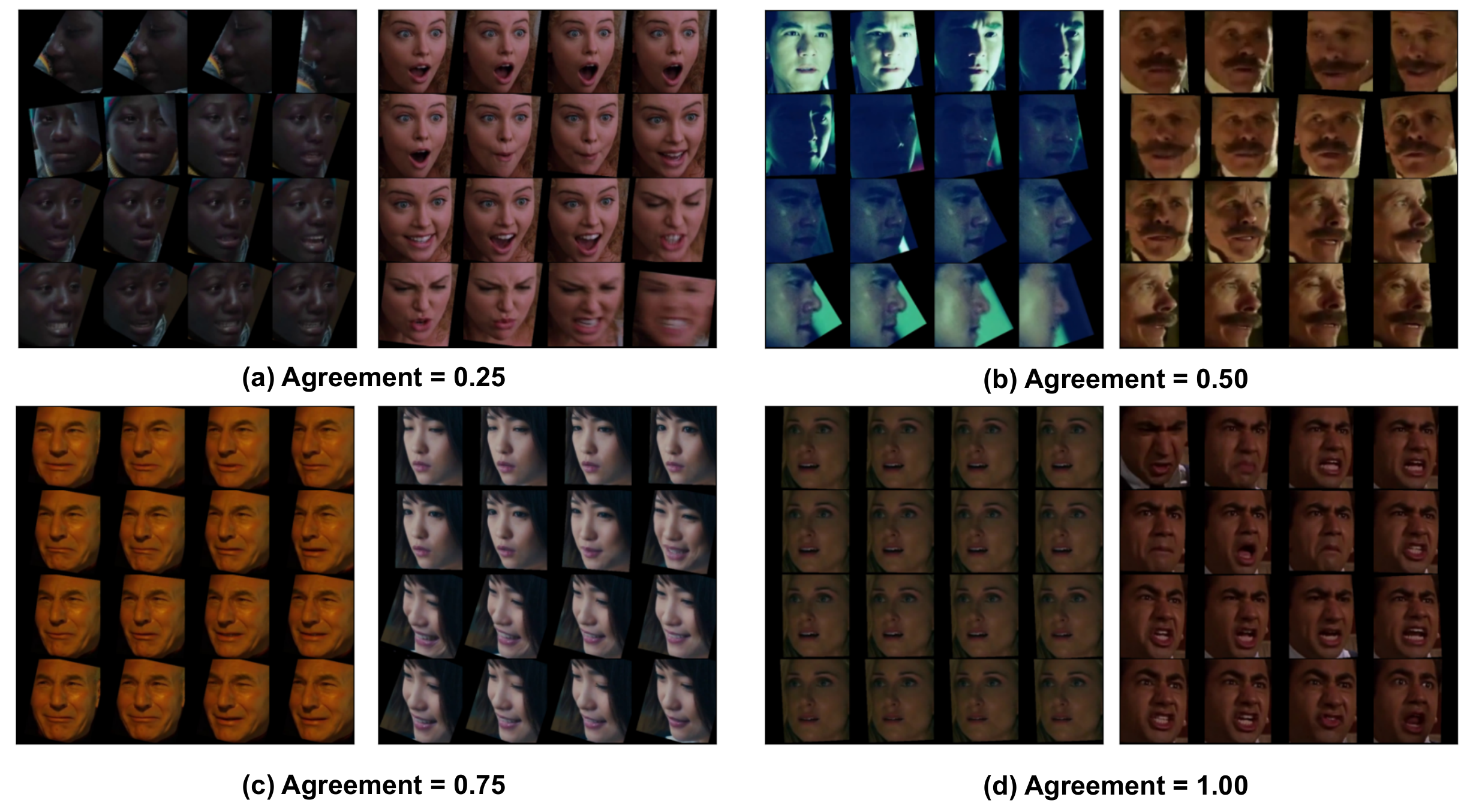}
  \caption{Sample visualization with different agreements.}
  \label{fig:sample}
\end{figure*}

\begin{figure*}[ht!]
  \centering
  \includegraphics[width=1\linewidth]{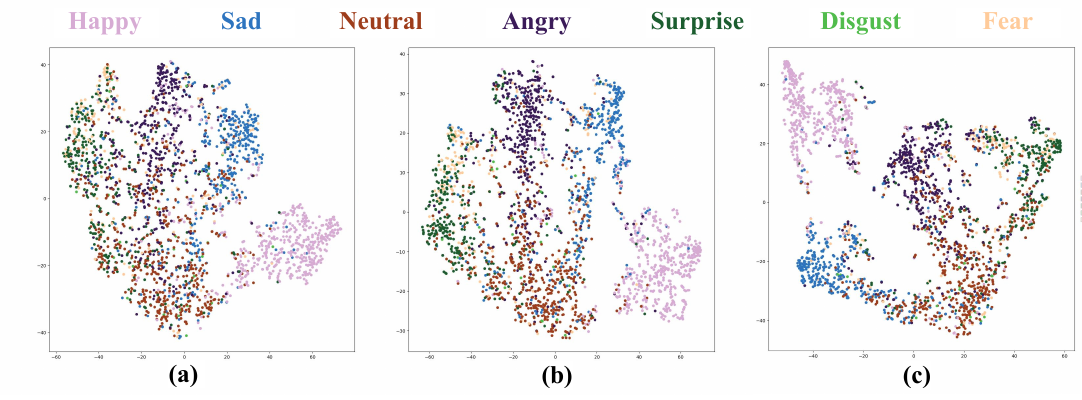}
  \caption{2D t-SNE visualization of dynamic facial expression features obtained with different agreements and expressions in table \ref{tab:ablat_agree_acc}. (a)Agreement = 0.5. (b)Agreement = 0.75. (c)Agreement = 1.00.}
  \label{fig:tsne}
\end{figure*}

\textbf{Sample Analysis.} Fig. \ref{fig:sample} displays visualizations of samples exhibiting varying degrees of agreement. In Fig. \ref{fig:sample}(a), four frames are grouped together to form one clip, resulting in four distinct predictions. The samples present a significant challenge, as the left sample exhibits notable variations in posture and the subject's dark complexion detracts from the clarity of facial characteristics, akin to shadowing, and the right sample showcasing multiple expressions such as \textit{Happy}, \textit{Surprise}, \textit{Angry}, and \textit{Fear}. Mislabeling such samples would have a limited impact on the model's performance. Samples in Fig. \ref{fig:sample}(b) exhibit significant changes in lighting and pose. Conversely, samples in Fig. \ref{fig:sample}(c) and (d) are more stable, with characters exhibiting minor changes or maintaining their expressions throughout. By evaluating the agreement of these samples, we can distinguish between hard and easy samples. This enables us to identify noisy easy samples, which have a more detrimental impact on the model, and to mine the hard samples as shown in Fig. \ref{fig:tsne}.

\begin{figure*}[ht!]
  \centering
  \includegraphics[width=1\linewidth]{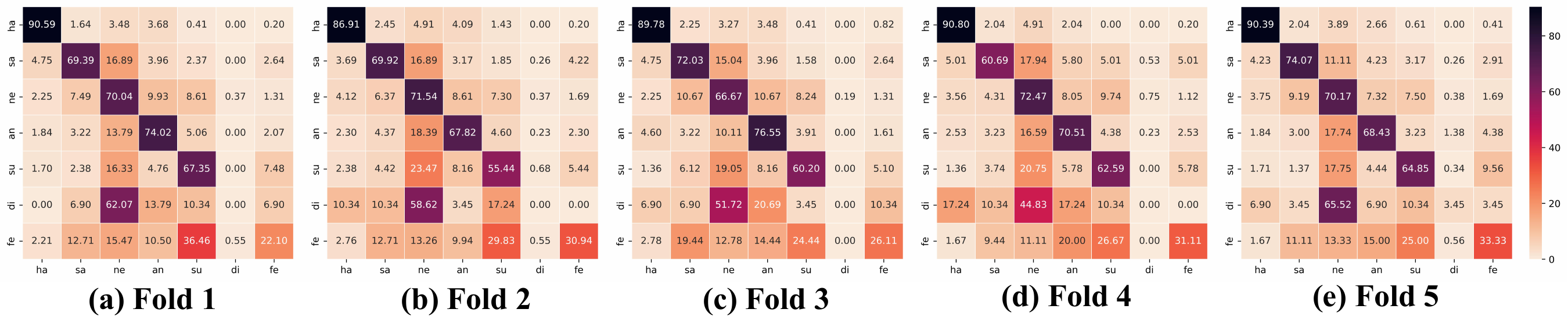}
  \caption{The confusion matrix of our proposed method evaluated on DFEW Fold 1-5. }
  \label{fig:confu}
\end{figure*}

\textbf{Confusion matrix.} To analyze the results of our proposed method evaluated on DFEW Fold 1\~5, we depict the confusion matrix. As shown in Fig. \ref{fig:confu}, our findings reveal that the predictive accuracy of the model for videos labeled as \textit{Disgust} is lower than other emotions, indicating the impact of label imbalance in DFEW. Specifically, the proportion of \textit{Disgust} videos is only 1.22\%, which results in the model's tendency to overlook these videos during training, thereby leading to inferior performance for this emotion. Moreover, a similar situation occurs for the \textit{Fear} label that has a proportion of 8.14\%. In this case, the model tends to misclassify some videos labeled as \textit{Fear} as other emotions due to the lack of sufficient training examples for this emotion. Furthermore, our analysis also reveals that the model tends to predict the label \textit{Neutral} more frequently, which is also supported by Fig. \ref{fig:tsne}.

Furthermore, the RDFER model demonstrates a propensity to predict the label as 'Neutral' on multiple occasions. This phenomenon may be attributable to the fact that Neutral expressions are more prevalent in daily life, or that the facial features of the dataset are more diverse, resulting in the model preferring Neutral as a prediction when confronted with samples that are challenging to evaluate. This prediction tendency may influence the model's precise recognition of other expression categories, particularly when the boundaries between these categories are more indistinct.

\section{Discussion}

Our approach is inspired by the Co-teaching series, which leverages model disagreement to update networks. However, Co-teaching requires training two networks simultaneously, resulting in a doubling of parameters and training time. Therefore, we investigate whether we can update a single network using the disagreement of one model towards different sampled clips of the video instead.

Decoupling \cite{malach2017decoupling} has demonstrated that samples with disagreement are valuable for learning. In the context of the DFER task, where facial movements vary across frames, our experiments have revealed that the model exhibits disagreement on different frames or clips within a single video. This characteristic implies that these samples are difficult to learn from due to their intricate facial movements and ambiguous expressions. Conversely, samples with high agreement are generally simpler. As the model confers high confidence to easy samples and low confidence to difficult ones, noisy-labeled easy samples carry more weight than noisy-labeled difficult samples. Thus, we define samples with high agreement and high loss as noisy samples that require less intensive learning, while samples with low agreement are considered difficult samples that necessitate more rigorous learning.

Our proposed approach for agreement presents a novel dimension to the current methods for handling noisy labels. While the existing methods rely on the small loss trick, our method has the added advantage of being able to differentiate between hard and noisy samples. Furthermore, it can be used alongside the small loss trick. Our simple implementation has yielded satisfactory results, and we posit that more complex frameworks can be employed to enhance the learning of hard and noisy samples. This could potentially replace the manual definition of these samples and the hyper-parameters of losses with established theoretical frameworks, thereby unlocking the full potential of the agreement approach. However, it must be noted that the approach of agreement requires the acquisition of outcomes from each frame in a given sample, thereby limiting its feasibility in tasks resembling DFER.

\section{Limitation}
\label{sec:limita}
Although RDFER did indeed achieve SOTA results, our current work is not without limitations that require further investigation.

The presence of hard and noisy samples represents a significant challenge that has been identified in our work, which warrants further investigation. Although notable results have been achieved by optimizing for these two sample types, there may still be other noisy samples, such as those caused by the inherent limitations of muscle motor units constrained by biological factors. Furthermore, additional statistical and systematic noise issues require further investigation.

Furthermore, the representational learning was enhanced by the separation of short-term movements from long-term mood changes, which resulted in an improvement in the effect. Nevertheless, a potential weak relationship between movements and mood changes may still impact the efficacy of representational learning.

In conclusion, although experiments were conducted on mainstream datasets with favorable outcomes, there is still a considerable distance to be traversed before real-world applications can be considered viable. Furthermore, larger samples and data are required to validate the effect of robust dynamic expression recognition. It is imperative that these limitations are addressed if dynamic expression recognition techniques and their applications targeting the field of computational visual media are to advance.

\section{Conclusion}
\label{sec:concl}

This article identifies the complex nature of the DFER problem, which involves hard and noisy samples that make the small loss trick ineffective. To address this issue, we propose a novel approach that involves discerning between hard and noisy samples using prediction agreement across various sample frames. This methodology strengthens hard sample learning while attenuating the learning of noisy samples. Additionally, we introduce a Key Expression Re-sampling Framework and a Dual-Stream Hierarchical Network to improve the model's representation learning ability and locate critical expressions in a video. We discuss the advantages and potential of the proposed agreement, and provide visualization of a set of samples with different agreements. Statistical analysis about the proportion and accuracy of the samples with different agreements and expressions are provided to discuss the property of agreement. Our experimental results on benchmark datasets demonstrate the superiority of our method over existing state-of-the-art approaches. 

\bibliographystyle{IEEEtran}
\bibliography{reference}


 




\vfill

\end{document}